\documentclass[conference]{IEEEtran}
\IEEEoverridecommandlockouts
\usepackage{cite}
\usepackage{amsmath,amssymb,amsfonts}
\usepackage{graphicx}
\usepackage{textcomp}
\usepackage{xcolor}
\def\BibTeX{{\rm B\kern-.05em{\sc i\kern-.025em b}\kern-.08em
    T\kern-.1667em\lower.7ex\hbox{E}\kern-.125emX}}

\usepackage{algorithm}
\usepackage{algorithmicx}
\usepackage[noend]{algpseudocode}% http://ctan.org/pkg/algorithmicx
\usepackage[caption=false,font=normalsize,labelfont=sf,textfont=sf]{subfig}
\usepackage{amsmath}
\usepackage[absolute]{textpos}
\usepackage{hyperref}
\usepackage{caption}

\newcommand{\refeq}[1]{Eq.~\eqref{#1}}

\DeclareMathOperator*{\argmin}{\arg\min}

\newcommand{\set}[1]{\mathcal{#1}}
\newcommand{\pnorm}[1]{\lVert{#1}\rVert}

\newcommand{\setX}{\ensuremath{\set{X}}}
\newcommand{\setY}{\ensuremath{\set{Y}}}
\newcommand{\x}{\ensuremath{\vec{x}}}

\newcommand{\y}{\ensuremath{y}}

\newcommand{\RN}{\mathbb{R}}

\newcommand{\autoencoder}{\ensuremath{\text{ae}}}
\newcommand{\encoder}{\ensuremath{\text{enc}}}
\newcommand{\decoder}{\ensuremath{\text{dec}}}
\newcommand{\loss}{\ensuremath{\ell}}
\newcommand{\transfer}{\ensuremath{f}}

\newcommand{\setD}{\ensuremath{\set{D}}}
\newcommand{\setDadapt}{\ensuremath{\set{D}_{*}}}
\newcommand{\E}{\mathbb{E}}

\newcommand{\refdef}[1]{Definition~\ref{#1}}
\newtheorem{definition}{Definition}
\newtheorem{remark}{Remark}

\usepackage{soul}

\begin{document}

% \begin{textblock}{12}(2,0.3)
% 	\noindent TODO
% \end{textblock}

\title{Unsupervised Unlearning of Concept Drift with Autoencoders
%\thanks{KM, CGP, MMP: This work has been supported by the European Research Council (ERC) under grant agreement No 951424 (Water-Futures), by the European Union’s Horizon 2020 research and innovation programme under grant agreement No 739551 (KIOS CoE), and from the Republic of Cyprus through the Deputy Ministry of Research, Innovation and Digital Policy.}
}
\author{
	\IEEEauthorblockN{
		Andr\'e Artelt\textsuperscript{a,b,c},
		Kleanthis Malialis\textsuperscript{b},
		Christos G. Panayiotou\textsuperscript{b, c},
		Marios M. Polycarpou\textsuperscript{b, c},
		Barbara Hammer\textsuperscript{a}
	}\\
	
	\IEEEauthorblockA{
    	\textsuperscript{a} \textit{Faculty of Technology}\\
    	\textit{Bielefeld University}, Bielefeld, Germany\\
    	Email: \{aartelt, bhammer\}@techfak.uni-bielefeld.de\\
    	ORCID: \{0000-0002-2426-3126, 0000-0002-0935-5591\}
	}\\
	
	\IEEEauthorblockA{
		\textsuperscript{b} \textit{KIOS Research and Innovation Center of Excellence}\\
		\textsuperscript{c} \textit{Department of Electrical and Computer Engineering}\\
		\textit{University of Cyprus}, Nicosia, Cyprus\\
		Email: \{malialis.kleanthis, christosp, mpolycar\}@ucy.ac.cy\\
		ORCID: \{0000-0003-3432-7434, 0000-0002-6476-9025, 0000-0001-6495-9171\}
	}\\
}

\maketitle

\begin{abstract}
Concept drift refers to a change in the data distribution affecting the data stream of future samples. % -- such non-stationary environments are often encountered in the real world.
Consequently, learning models operating on the data stream might become obsolete, and need costly and difficult adjustments such as retraining or adaptation. Existing methods usually implement a local concept drift adaptation scheme, where either incremental learning of the models is used, or the models are completely retrained when a drift detection mechanism triggers an alarm. This paper proposes an alternative approach in which an unsupervised and model-agnostic concept drift adaptation method at the global level is introduced, based on autoencoders. Specifically, the proposed method aims to ``unlearn'' the concept drift without having to retrain or adapt any of the learning models operating on the data. An extensive experimental evaluation is conducted in two application domains. We consider a realistic water distribution network with more than 30 models in-place, from which we create 200 simulated data sets / scenarios. We further consider an image-related task to demonstrate the effectiveness of our method.
\end{abstract}


\section{Introduction}\label{sec:intro}
Concept drift is one of the major challenges encountered by learning algorithms in data stream mining \cite{ditzler2015learning}. It refers to the problem of dealing with a non-stationary data distribution that evolves over time as data continually arrive in a streaming manner. If not properly addressed, learning models may become obsolete with severe consequences in practical applications.

Consider, for example, a domain area in which multiple learning models have been trained, each specialised for a particular downstream task; i.e., a set of (different) models operate on the same data (stream). For instance, in water distribution networks~\cite{kyriakides2014intelligent} each model corresponds to a different downstream task, such as predicting water pressure at a node from neighboring nodes with pressure sensors installed, detecting water leakage, and identifying the leakage location (i.e. isolation). Different types of models might have been trained, i.e., regression (pressure prediction) and classification (leakage detection and isolation) models, as well as different learning paradigms might have been used, i.e., supervised learning (pressure prediction) and unsupervised learning / anomaly detection (leakage detection).

In the presence of concept drift, most - if not all - downstream models will be incapable of adequately performing their corresponding tasks. For instance, in water distribution networks examples of concept drift constitute changes in demands, and sensor faults. As a result, the traditional approach to be followed would be to update (either by incremental learning or complete re-training) each downstream model to maintain an optimal performance. Numerous methods~\cite{ditzler2015learning, gama2014survey} have been proposed for concept drift adaptation, which we review later in Section~\ref{sec:related}. Indeed, for some types of drift, e.g. water demand changes, the traditional approach should rightfully be applied. However, for some other drift types, e.g. sensor faults, existing methods become impractical, costly, and fail to scale well when the number of models is large.
% \begin{figure}[t]
%  	\centering 	
%  	\includegraphics[scale=0.5]{images/illustration.pdf}
%  	\caption{Downstream learning models applied to a data stream.}
%         \label{fig:motivation}
% \end{figure}

The contributions made in this work are the following:
\begin{itemize}
    \item We propose an unsupervised learning method for global concept drift adaptation based on autoencoders. Our method aims to revert the data distribution to the state it was before the concept drift had occurred, without the need to modify or even consider any of the existing downstream models -- i.e. our proposed method is completely model-agnostic. Its key advantage is that no ground truth information (e.g., labels in a classification task) is required from human experts.
    
    \item We conduct empirical evaluations, involving a diverse set of different domains and scenarios, and demonstrate the effectiveness of our proposed method. Specifically, we evaluate our method in $200$ simulated data sets / scenarios from a realistic water distribution network (each corresponding to different drift characteristics) with more than $30$ downstream models. Furthermore, evaluation is performed in an image-related task as well.
\end{itemize}

The remainder of this paper is organized as follows: First (Section~\ref{sec:problem}), we formally introduce the problem of local and global concept adaptation which we are dealing with in this work. Next (Section~\ref{sec:related}), we present related and existing work to place our work in the literature. In Section~\ref{sec:method} we introduce and describe our proposed method for implementing a global concept drift adaptation using autoencoders, which we empirically evaluate in several case studies (Section~\ref{sec:exp}). Section~\ref{sec:discussion} discusses important remarks and characteristics of the proposed method, before Section~\ref{sec:conclusion} concludes this work including some pointers to future directions.

\section{Concept Drift Adaptation Problem}\label{sec:problem}
In this work we deal with the problem of adapting a set of models to concept drift -- because in a supervised scenario concept drift becomes visible by means of a drop in accuracy or increasing loss~\cite{gama2014survey}, we formalize the concept drift adaptation problem as follows:
\begin{definition}[Local Concept Drift Adaptation Problem]\label{def:conceptdriftadaptation}
We are given a set models $m_i:\setX\to\setY$ which were trained on data from the distribution $p_{\setX,\setY}$.
Next, we assume that some type of concept drift happens, and the data distribution changes to $p'_{\setX,\setY}$.
We assume that this concept drift affects the performance of some (possibly all) models $m_i(\cdot)$:
\begin{equation}\label{eq:concepdrift:modelaccuracy}
    \underset{x,y\sim p'_{\setX,\setY}}{\E}\left[\loss(m_i,x,y)\right] > \underset{x,y\sim p_{\setX,\setY}}{\E}\left[\loss(m_i,x,y)\right]
\end{equation}
where $x,y\sim p_{\setX,\setY}$ denotes samples from the joint distribution of $\setX,\setY$, and $\loss(\cdot)\mapsto\RN_{+}$ denotes a suitable function for measuring the performance of the models (e.g., mean-squared error).

The \textit{Local Concept Drift Adaptation Problem} is to adapt the models $m_i(\cdot)$, yielding models $m'_i(\cdot)$, to the changed distribution such that the performances increases:
\begin{equation}
    \underset{x,y\sim p'_{\setX,\setY}}{\E}\left[\loss(m_i,x,y)\right] > \underset{x,y\sim p'_{\setX,\setY}}{\E}\left[\loss(m_i',x,y)\right]
\end{equation}
\end{definition}
In this work, we propose an alternative to adapting the models $m_i(\cdot)$ directly -- i.e. adapting ``locally'' to the concept drift --, as it is done in~\refdef{def:conceptdriftadaptation}. Instead of a local adaptation, we aim for a global adaptation of the data distribution that is agnostic of any used models -- i.e. we want to transform the observed data $\x\in\setX$ such that the influence of the concept drift is removed. We formalize this as follows:
\begin{definition}[Global Concept Drift Adaptation Problem]\label{def:conceptdriftadaptation:global}
%We consider the same setting as in~\refeq{eq:concepdrift:modelaccuracy}: That is, we are given a set models $m_i:\setX\to\setY$ which were trained on data from distribution $p_{\setX,\setY}$, and we assume that some type of concept drift happens and the data distribution changes to $p'_{\setX,\setY}$.
%We assume that this concept drift (i.e. change of the data distribution) affects the performance of some (possibly all) models $m_i(\cdot)$ -- see~\refeq{eq:concepdrift:modelaccuracy}.

The \textit{Global Concept Drift Adaptation Problem} is to adapt to the changed distribution by transforming the samples $\x\in\setX$ using a mapping $f:\setX\to\setX$, such that the performances of $m_i(\cdot)$ increases:
\begin{equation}
    \underset{x,y\sim p'_{\setX,\setY}}{\E}\left[\loss(m_i,x,y)\right] > \underset{x,y\sim p'_{\setX,\setY}}{\E}\left[\loss(m_i,f(x),y)\right]
\end{equation}
\end{definition}

\begin{remark}
Note that the key difference between~\refdef{def:conceptdriftadaptation} and~\refdef{def:conceptdriftadaptation:global} is the way the concept drift adaptation is done. In the former each model $m_i(\cdot)$ is adapted separately, while in the latter a global adaptation of the observed data is computed. Local and global adaption complement each other; local adaption can be appropriate for some drift types (e.g., water network expansion, water demand changes) while global adaption is appropriate for other types (e.g., sensor faults). \refdef{def:conceptdriftadaptation} proposes to adapt/change each model while~\refdef{def:conceptdriftadaptation:global} does not change any model at all.
\end{remark}

\paragraph*{Our Contribution}
This work proposes a method for a global drift adaptation (\refdef{def:conceptdriftadaptation:global}) which is completely model-agnostic and unsupervised -- i.e. we do not make any assumptions on the models $m_i(\cdot)$ and do not assume, in contrast to many other methods, that we observe labeled data.

\section{Related Work}\label{sec:related}
Existing methods on addressing concept drift (\refdef{def:conceptdriftadaptation} and~\refdef{def:conceptdriftadaptation:global}) can be (loosely) categorised into the following three categories~\cite{ditzler2015learning, gama2014survey}: concept drift adaptation, transfer learning, and representation learning.

\subsubsection{\textbf{Concept Drift Adaptation}}
Methods in this category are further grouped as passive, active, and hybrid methods.

\textbf{Passive methods}.
% \subsubsection{\textbf{Passive} methods}
These methods implicitly adapt to concept drift by using incremental learning, that is, they continually update a learning model \cite{losing2018incremental} -- these methods constitute a solution to the concept drift adaptation problem (\refdef{def:conceptdriftadaptation}) since they adapt each model $m_i(\cdot)$. Passive methods make use of memory components (also referred to as memory-based methods), such as a moving window that maintains a list of the most recent examples. Memory-based methods have the disadvantage of having to pre-determine the ``right'' memory size, therefore, mechanisms to resolve this have been introduced, such as having an adaptive window \cite{widmer1996learning}. Other passive methods use ensembling where a set of classifiers dynamically grows or shrinks to maintain good performance, such as the Learn++.NSE method \cite{elwell2011incremental}.%, and DDD \cite{gama2004learning} which maintains ensembles with different diversity levels, as a different level of ensemble diversity may be more appropriate before and after a drift for better generalisation.

Besides the assumption of observing labeled data, another key challenge is class imbalance \cite{wang2018systematic} and to address it, incremental learning has been used in combination with different mechanisms, e.g., memory-based models~\cite{JakobAHH22} like having one memory component per class \cite{malialis2018queue}, (adaptive) rebalancing \cite{malialis2021online,vaquet2020sam}, and bagging \cite{wang2014resampling, cano2022rose}.

Furthermore, most existing methods assume supervision. Incremental learning has been used in conjunction with other learning paradigms, such as active learning \cite{malialis2022nonstationary, zliobaite2013active, malialis2022data} and unsupervised learning \cite{li2023}.

% \subsubsection{\textbf{Active} methods}
\textbf{Active methods}.
These methods use an explicit mechanism for concept detection and are, typically, referred to as change detection-based methods. Some methods use statistical tests, such as independence tests~\cite{li2023, pmlr-v119-hinder20a}, or JIT classifiers \cite{alippi2008justI}) which propose two CUSUM-inspired tests capable of detecting abrupt and smooth changes. Some others use a threshold-based mechanism, e.g., DDM \cite{gama2004learning} which uses one threshold value to raise a warning flag and another to trigger a drift alarm.

In contrast to passive methods which incrementally update a learning model, active methods discard the existing model and create a new one that performs a complete retraining as soon as a concept drift alarm is triggered -- therefore, similar to passive methods, these can be interpreted as a solution to the concept drift adaptation problem (\refdef{def:conceptdriftadaptation}). Moreover, \textbf{ensembling} has also been used in active methods \cite{krawczyk2017ensemble}.

% \subsubsection{\textbf{Hybrid}}
\textbf{Hybrid methods}.
Hybrid methods combine the advantages of both approaches, for instance, HAREBA \cite{malialis2022hybrid} combines a threshold-based drift detection mechanism with adaptive rebalancing to cope with class imbalance, \cite{zliobaite2013active} uses explicit drift detection with incremental active learning, while \cite{li2023} proposes strAEm++DD, an autoencoder-based incremental learning method with drift detection.

\subsubsection{\textbf{Transfer Learning}}
Transfer learning~\cite{ZhuangQDXZZXH21,pan2010survey,weiss2016survey} can also be used for adapting to a changed distribution -- i.e. concept drift adaptation. Unfortunately, many transfer learning methods require labeled data~\cite{weiss2016survey}. A few unsupervised methods, however, exist: For instance~\cite{VaquetMSH22,yao2020unsupervised,sanodiya2020unsupervised}.

\subsubsection{\textbf{Representation Learning}}
Representation learning~\cite{bengio2013representation} aims to learn a data representation that can be used by different learning models operating on the same data domain -- e.g. learning models operating on text data might work on the same text embedding (i.e. text representation). However, such learned representations are also affected by concept drift and therefore have to be adapted and/or be as robust and invariant as possible. Only very little work on adapting representations exist: For instance, in~\cite{sokar2021learning} a method for learning invariant representations for continual learning (a very special case where new tasks must be learned over time) is proposed.

\section{Unsupervised Unlearning of Concept Drift with Autoencoders}\label{sec:method}
Our proposed method is unsupervised, makes no assumption on the models $m_i(\cdot)$, and consists of two major parts:
\begin{enumerate}
    \item \textbf{Distribution Learning:} We build an autoencoder that attempts to learn the data distribution so that we can characterize the influence of concept drift on observed samples.
    We note that the proposed method is agnostic to the drift detection mechanism used. Our focus is not on the problem of concept drift detection, but rather assumes that an effective and efficient concept drift detection mechanism is in place and that the influence of concept drift can be characterized by the autoencoder.
    
    \item \textbf{Unsupervised Transfer Learning:} When concept drift is detected, we build a function that tries to ``undo'' the concept drift -- i.e. implementing $f(\cdot)$ from~\refdef{def:conceptdriftadaptation:global}. That is, we try to transform samples observed from the drifted data distribution, appear like they had been observed from the original (pre-drift) data distribution, by using the autoencoder built in the beginning.
\end{enumerate}
Our entire methodology is described in detail in Algorithm~\ref{alg:method}.

\subsection{Autoencoder for Distribution Learning}
Concept drift refers to a change in the data distribution $p_{\setX,\setY}$ which might affect downstream models $m(\cdot)$ that operate on the data from the domain $\setX$. However, since the true distribution $p_{\setX,\setY}$ is usually not known, it is estimated from the data. Having an estimate of $p_{\setX,\setY}$ enables one not only to detect concept drift~\cite{gama2014survey} but also to distinguish between samples before/after the concept drift -- this, for instance, can be used to learn something about the concept drift itself~\cite{hinder2022expldrift}.

We use an autoencoder $\autoencoder: \setX \to \setX$ for modeling the data distribution $p_\setX$. This autoencoder $\autoencoder(\cdot)$ consists of an encoder $\encoder: \setX \to \setX'$ and a decoder $\decoder: \setX' \to \setX$, whereby $\setX'$ denotes the space of the encoding:
\begin{equation}\label{eq:autoencoder}
\autoencoder: x \mapsto (\decoder \circ \encoder)(x)
\end{equation}
Learning an autoencoder~\refeq{eq:autoencoder} from a given data set $\setD \subset \setX$ means finding an encoder $\encoder(\cdot)$ and a decoder $\decoder(\cdot)$ such that the reconstruction loss $\loss(\cdot)$\footnote{E.g. mean-squared error or some other function for penalizing differences between two samples.} is minimized:
\begin{equation}\label{eq:autoencoder:learning}
\underset{\encoder(\cdot),\,\decoder(\cdot)}{\inf}\,\sum_{x_i\,\in\,\setD}\loss\big(\underbrace{(\decoder \circ \encoder)(x_i)}_{\hat{x}_i}, x_i\big)
\end{equation}

In this work, both the encoder $\encoder(\cdot)$ and decoder $\decoder(\cdot)$ are parameterized functions and the problem of learning the autoencoder~\refeq{eq:autoencoder:learning} can be rewritten as an optimization problem over the parameters $\theta_1\in\RN^m$ and $\theta_2\in\RN^m$:
\begin{equation}\label{eq:autoencoder:learning_final}
    \underset{\theta_1,\,\theta_2\in\RN^m}{\argmin}\,\sum_{x_i\,\in\,\setD}\loss\big(\underbrace{(\decoder_{\theta_2} \circ \encoder_{\theta_1})(x_i)}_{\hat{x}_i}, x_i\big)
\end{equation}

\subsection{Unsupervised Transfer Learning using Autoencoders}\label{sec:undoing_change}
We assume that the data distribution $p_\setX$ changes at some point in time, which then leads to a larger reconstruction error of the autoencoder~\refeq{eq:autoencoder}. While using the autoencoder to discriminate between samples from before and after the drift could be an obvious choice as in \cite{JaworskiRA20}, as mentioned, the proposed approach is agnostic to the concept drift detection mechanism used.

In order to ``unlearn'' the concept drift, we implement the global adaptation $f(\cdot)$ from~\refdef{def:conceptdriftadaptation:global} by learning a mapping $\transfer: \setX \to \setX$ such that the reconstruction loss becomes smaller again -- we use the reconstruction error as a proxy for how well a given sample fits to the original data distribution. We infer this mapping $\transfer(\cdot)$ from a given data set of unlabeled samples $\setDadapt \subset \setX$ under the new distribution (i.e. some samples observed after the concept drift).

%Finding such a function $\transfer(\cdot)$ is modeled as the following optimization problem:
%\begin{equation}
%\underset{\transfer\,\in\,\setTransfer}\inf\;\sum_{x_j\,\in\,\setDadapt}\loss\big(\underbrace{(\autoencoder \circ \transfer)(x_j)}_{\hat{x}_j}, \transfer(x_j)\big)
%\end{equation}
%or equivalently (explicitly using the encoder and decoder):
%\begin{equation}
%\underset{\transfer\,\in\,\setTransfer}\inf\;\sum_{x_j\,\in\,\setDadapt}\loss\big(\underbrace{(\underbrace{\decoder \circ \encoder}_{\text{Autoencoder}} \circ \transfer)(x_j)}_{\hat{x}_j}, \transfer(x_j)\big)
%\end{equation}
%\textcolor{red}{What if reconstruction loss is still small and can not be improved}
%\textcolor{red}{$\transfer(\cdot)$ and autoencoder could be chained together -- con: undoing becomes computationally more expensive because of large MLP, might be problematic for real-time applications with a high sampling rate.}
In this work we parameterize the mapping $\transfer(\cdot)$, and optimize over the parameters $\theta$ in order to find the final mapping $\transfer(\cdot)$:
\begin{equation}\label{eq:fitting_transform_opt}
\underset{\theta}\argmin\;\frac{1}{|\setDadapt|}\sum_{x_j\,\in\,\setDadapt}\loss\big(\underbrace{(\autoencoder \circ \transfer_{\theta})(x_j)}_{\hat{x}_j}, \transfer_{\theta}(x_j)\big)
\end{equation}
%or equivalently:
%\begin{equation}
%\underset{\theta}\argmin\;\sum_{x_j\,\in\,\setDadapt}\loss\big(\underbrace{(\decoder \circ \encoder \circ \transfer_{\theta})(x_j)}_{\hat{x}_j}, \transfer_{\theta}(x_j)\big)
%\end{equation}
Note that it would also be possible to chain $\transfer(\cdot)$ and the autoencoder together, however in this case the final function (i.e. $\autoencoder \circ f$) for adapting to the drift would be computationally more complex and might therefore be less suited for real-time scenarios and scenarios where adapting to drift must be performed on devices with limited hardware capabilities (i.e. on-device learning).

In order to avoid less useful solutions -- e.g. the mapping $\transfer(\cdot)$ might yield the same output or might exploit some special particularities of the autoencoder, we propose to add the $l_1$ regularization term $C\cdot\pnorm{\transfer_{\theta}(x_j) - x_j}_1$ to the optimization problem~\refeq{eq:fitting_transform_opt}:
\begin{equation}\label{eq:fitting_transform_opt:final}
\begin{split}
\underset{\theta}\argmin\;&\frac{1}{|\setDadapt|}\sum_{x_j\,\in\,\setDadapt}\loss\big(\underbrace{(\autoencoder \circ \transfer_{\theta})(x_j)}_{\hat{x}_j}, \transfer_{\theta}(x_j)\big) +\\&\quad C\cdot\pnorm{\transfer_{\theta}(x_j) - x_j}_1
\end{split}
\end{equation}
where $C\geq 0$ denotes regularization strengths which allow balancing between the different terms in the objective. The reasoning behind this regularisation is to minimize the number of features that are changed by $\transfer(\cdot)$.

%\subsection{Unsupervised Unlearning of Concept Drift}\label{sec:unlearningdrift}
As pointed out in the beginning of Section~\ref{sec:method}, we can use the unsupervised transfer learning using autoencoder from Section~\ref{sec:undoing_change} as a method for globally adapting (i.e. unlearning) to concept drift as stated in~\refdef{def:conceptdriftadaptation:global} -- i.e. removing the influence of concept drift on the data space such that all models working on the data space are still applicable and perform reasonable well -- see Algorithm~\ref{alg:method}.
\begin{algorithm}[t!]
	\caption{Unsupervised Unlearning of Concept Drift}
	\label{alg:method}
	\begin{algorithmic}[1]
		
		\Statex \textbf{Arguments:} Data stream of unlabeled samples $(\x_t, \x_{t+1}, \dots)$; A set of already trained downstream tasks $\{m_i(\cdot)\}$
		\State Solve~\refeq{eq:autoencoder:learning_final} \Comment{Build the autoencoder on a batch $\setD$ of data before any concept drift happens.}
		
		\Statex \textbf{Main:}
		  \If{d(t)==True}     \Comment{Test for concept drift using some drift detection method $d(\cdot)$.}
                \State Collect $\setD_{*}$  \Comment{Build a data set for ``undoing'' the drift.}
                \State Solve~\refeq{eq:fitting_transform_opt:final}     \Comment{Build $\transfer(\cdot)$ for ``undoing'' the drift.}
            \Else
                \State $\y_i = m_i(\transfer(\x_t)) \quad \forall\,i$    \Comment{Apply current sample $\x_t$ to $\transfer(\cdot)$ and then to the downstream tasks $\{m_i(\cdot)\}$.}
            \EndIf
	\end{algorithmic}
\end{algorithm}

\section{Experiments}\label{sec:exp}
We empirically evaluate the performance of our proposed methodology for global concept drift adaptation on a set of diverse data sets. For technical implementation details of the following experiments, we refer to the Python implementation which is publicly available on GitHub\footnote{https://github.com/HammerLabML/\\UnsupervisedUnlearningConceptDriftAutoencoders}. 

\subsection{Data}\label{sec:exp_data}
We consider two data sets for our experiments:
\subsubsection{Digits}
\begin{figure}
    \centering

	\begin{minipage}[b]{0.49\textwidth}
		\includegraphics[width=\textwidth]{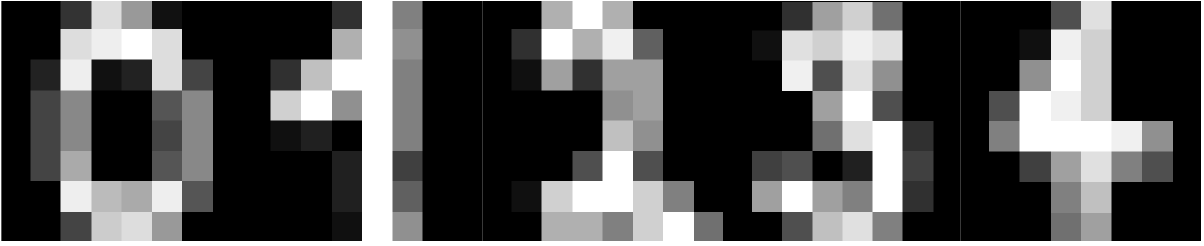}  
		\caption*{Original samples} 
	\end{minipage}
        \hfil
 	\begin{minipage}[b]{0.49\textwidth}
		\includegraphics[width=\textwidth]{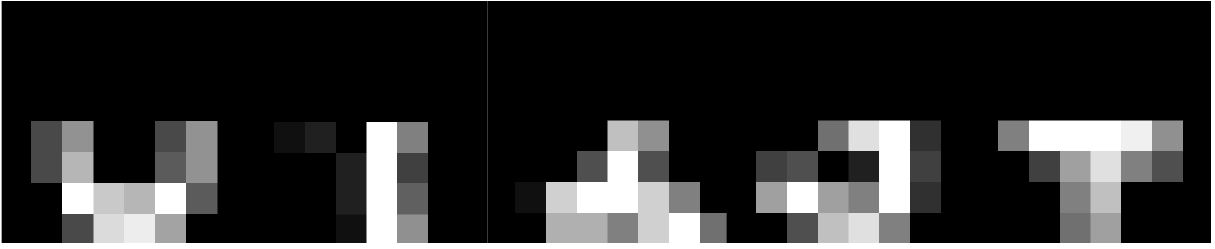}  
		\caption*{Samples after concept drift} 
	\end{minipage}
    \caption{Illustration of concept drift on the digits data set.}
    \label{fig:digits}
\end{figure}
We use the first five digits (i.e. 0-4) from~\cite{ocr} as a data set and build a digit classifier (this is the \textit{single} downstream model $m(\cdot)$) by using logistic regression. We introduce concept drift in the test data by setting all pixels in the upper half of the image to zero -- see Fig.~\ref{fig:digits} for an illustration. We use a $10$-fold cross-validation to get statistically meaningful results.
For a comparison, we also completely retrain the classifier using a labeled training set -- by this estimate how well a supervised drift adaptation could be, assuming labeled would be available.

\subsubsection{Hanoi}

\begin{figure}
    \centering
    \includegraphics[scale=0.45]{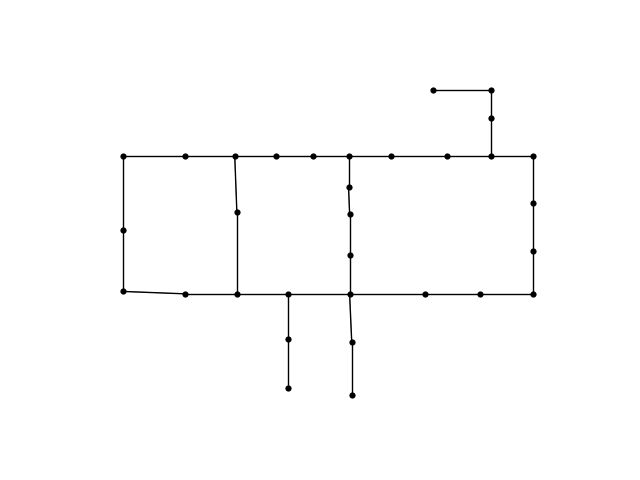}
    \caption{Hanoi water distribution network~\cite{vrachimis2018leakdb}.}
    \label{fig:hanoi}
\end{figure}

In this experiment, we use the Hanoi water distribution network~\cite{vrachimis2018leakdb}, shown in Figure~\ref{fig:hanoi}. We place a pressure sensor at every node in the network (there are $32$ nodes in total) and build a virtual sensor for nearly every sensor%\footnote{We exclude the first node because its pressure is always zero.}
-- i.e. we have $31$ downstream regression tasks $m(\cdot)$ that try to predict the pressure at a particular (different) location.

In order to get statistically meaningful results, we generated a total number of $200$ scenarios  
For each scenario, we introduce concept drift by selecting a random pressure sensor to become faulty -- i.e. placing a random sensor fault (with random parameters) somewhere in the network. To simulate real-world characteristics, we consider the following types of sensor faults~\cite{Reppa2016}:
\begin{enumerate}
    \item A constant offset of the sensor's measurement compared to the true measured quantity.
    \item Gaussian noise added to the sensor measurements.
    \item Sensor power failures, which result in the measurement being equal to zero.
    \item An offset of the sensor measurement, linearly proportional to the true measured quantity.
\end{enumerate}

\subsection{General Setup}\label{sec:exp_setup}
The general setup of the experiments is the same for all data sets:
\begin{enumerate}
    \item Train the downstream task models $m(\cdot)$ as well as the autoencoder $\autoencoder(\cdot)$ on a time window without any concept drift.
    \item Evaluate the downstream task models $m(\cdot)$ on the remaining samples before the concept drift occurs -- and also on samples after the concept drift occurred for evaluating the influence of the concept drift on the downstream tasks.

    Note that, for the purpose of evaluating our methodology, we do not build any concept drift detector but rather use the available ground truth.
    \item Build the unsupervised global concept drift adaptation function on a small time window after the concept drift occurred.
    \item Evaluate the downstream task models $m(\cdot)$ after applying the undoing function $\transfer(\cdot)$ on the samples after the concept drift.
    \item As a baseline, we use the autoencoder $\autoencoder(\cdot)$ as an implementation of $\transfer(\cdot)$ -- i.e. we evaluate if and how well the reconstruction of the autoencoder is already able to adapt the concept drift.
\end{enumerate}

In both cases, we implement the autoencoder as a multi-layer perceptron (MLP, i.e., a standard fully-connected feed-forward neural network) and the global concept drift adaptation function $\transfer(\cdot)$ is implemented as an affine mapping.

\subsection{Results}\label{sec:exp_results}
\begin{figure}
    \centering
    \includegraphics[scale=0.2]{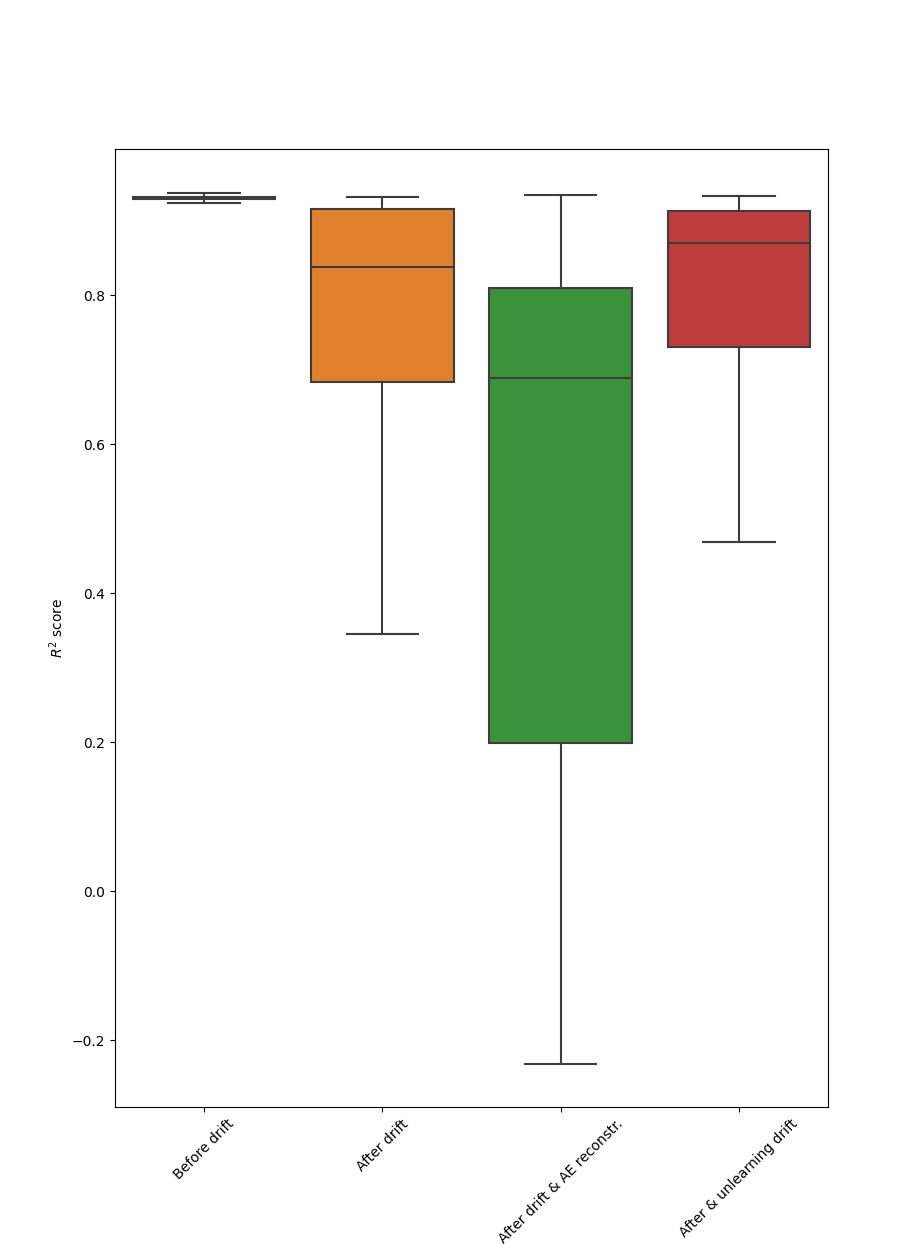}
    \caption{Results on the Hanoi data set for downstream task $4$ -- we plot the $R^2$ scores over all scenarios.}
    \label{fig:exp:hanoi:results:sample}
\end{figure}

\begin{table*}
\caption{Results on the digits data set -- we report the mean and variance over all ten folds, all numbers are rounded to two decimal points.}
\centering
\footnotesize
\begin{tabular}{|c||c|c||c|c||c||}
 \hline
 \textit{Accuracy$\uparrow$} & Before drift & After drift & After drift \& AE reconstruction & After \& drift adaptation & Supervised retraining  \\
 \hline\hline
  & $0.98 \pm 0.02$ & $0.68 \pm 0.05$ & $0.74 \pm 0.07$ & $0.77 \pm 0.08$ & $0.93 \pm 0.02$ \\
 \hline
 \hline
\end{tabular}
\label{table:exp:results:digits}
\end{table*}
The results of the experiments on the digits data set are shown in Table~\ref{table:exp:results:digits}.
For the Hanoi data set, we show the results for each downstream task and over all scenarios in Figure~\ref{fig:exp:hanoi:results} -- the detailed numbers are given as in Table~\ref{table:exp:results:hanoi} in the appendix. Note that, in order to get rid of outliers distorting the box plots, we filtered out all scenarios where the reconstruction of the autoencoder could not be improved, which leaves us with $138$ scenarios left. Furthermore, since there is a lot of information in Figure~\ref{fig:exp:hanoi:results}, we also provide the results of a single downstream task in Figure~\ref{fig:exp:hanoi:results:sample} for illustrative purposes.
\begin{figure*}[t!]
    %\centering
        \caption{Results on the Hanoi data set -- we plot the $R^2$ scores for each of $31$ downstream tasks over all scenarios.}
	\begin{minipage}[b]{0.49\textwidth}
		\includegraphics[width=\textwidth]{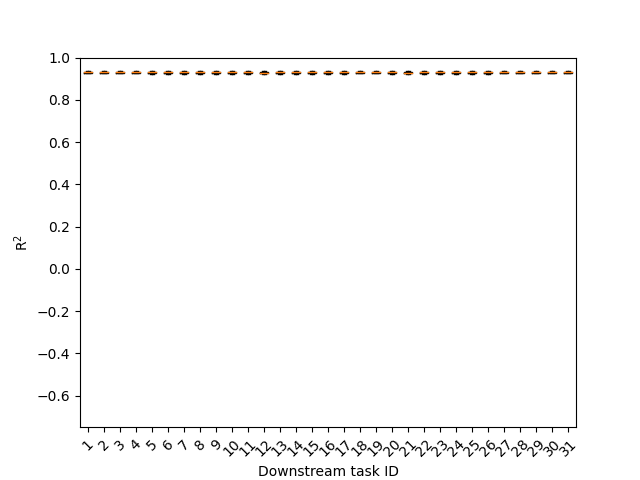}  
		\caption*{Before drift} 
	\end{minipage}
	\hfill\allowbreak%
 	\begin{minipage}[b]{0.49\textwidth}
		\includegraphics[width=\textwidth]{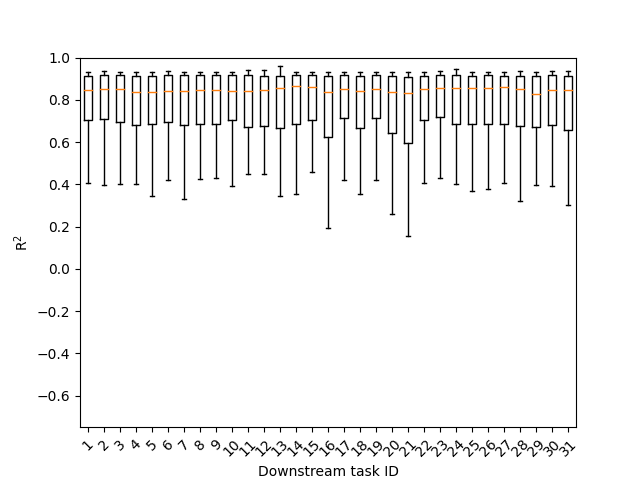}  
		\caption*{After drift} 
	\end{minipage}
        \hfill\allowbreak%
 	\begin{minipage}[b]{0.49\textwidth}
		\includegraphics[width=\textwidth]{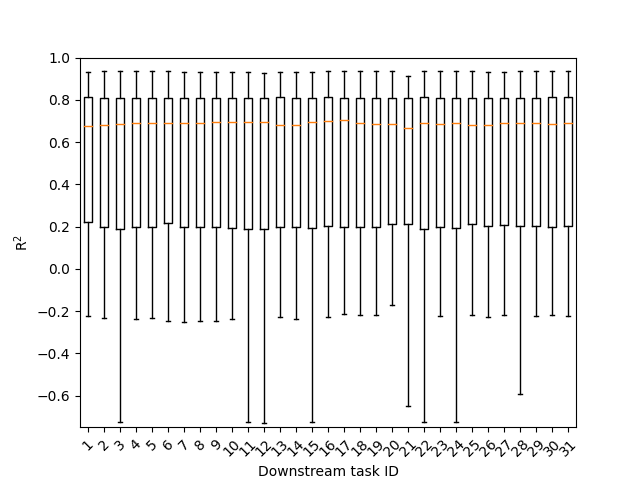}  
		\caption*{After drift, baseline method (AE reconstr.)} 
  	\end{minipage}
  	\hfill\allowbreak%
  	\begin{minipage}[b]{0.49\textwidth}
		\includegraphics[width=\textwidth]{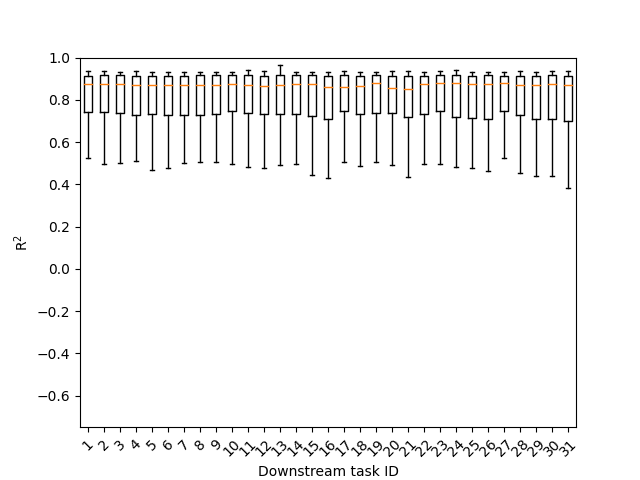}  
		\caption*{After drift, proposed method (unlearning drift)} 
	\end{minipage}
    \label{fig:exp:hanoi:results}
\end{figure*}

In both cases (data sets), we observe a significant improvement of the downstream task models $m(\cdot)$ after applying $\transfer(\cdot)$ to the drifted data. We also observe a significant improvement of the baseline where we use the autoencoder reconstruction instead of $\transfer(\cdot)$ for undoing the drift. We observe that supervised retraining (assuming labels are available) would lead to a significant performance boost but still a bit worth than the performance before the drift which indicates that the new concept is more difficult to learn. In the case of the Hanoi data set, we also observe a significant improvement in the variance -- i.e. solutions are more stable.
These findings demonstrate a strong performance of our proposed method for global concept drift adaptation.

\section{Discussion}\label{sec:discussion}
%\noindent \textbf{Advantages.} The key advantage of our proposed method is that it works completely unsupervised -- i.e. no labeled samples are required, which in the real world are often not available or are very costly to obtain -- and it is completely agnostic of the number and types of downstream tasks and models $m_i(\cdot)$, which becomes handy in case of many ``complex'' downstream tasks that require a large amount of (labeled) data for training.

\textbf{Computational aspects.} The computational complexity of our proposed method mainly depends on the implementation of $\transfer(\cdot)$ because the autoencoder, which might be a rather complex neural network, is only trained once in the beginning and is not changed anymore afterwards. The complexity of $\transfer(\cdot)$ (e.g. size of the neural networks that implements $\transfer(\cdot)$) then determines how many samples are needed for building $\transfer(\cdot)$ and consequently how much time this process of building $\transfer(\cdot)$ takes -- here time not only refers to the training time of $\transfer(\cdot)$ but also to the time until we can adapt to the concept drift and can continue using the downstream task models $m(\cdot)$. However, in this work, we observed that often a relatively simple architecture of $\transfer(\cdot)$ is already sufficient for adapting to the concept drift, and therefore only a small set of samples is needed which leads to a fast training of $\transfer(\cdot)$ as well.

\textbf{Limitations.} A potential limitation of our proposed approach is the choice of the autoencoder itself for distribution learning. If the concept drift does not lead to a larger reconstruction loss of the autoencoder, our proposed method is not able to learn a function for adapting the drift. We, therefore, suggest applying our proposed methodology only in cases where the concept drift manifests itself in a significantly large reconstruction loss of the autoencoder.

\section{Conclusion and Future Work}\label{sec:conclusion}
In this work, we proposed an unsupervised methodology for global concept drift adaptation using autoencoders -- i.e. ``undoing'' concept drift in an unsupervised manner, which has the advantage that no downstream model $m_i(\cdot)$ must be retrained or adapted for which labeled samples would be required. We empirically evaluated our proposed method on several scenarios from different domains and observed a strong performance of our proposed method.

Based on this initial work, a couple of potential directions for future research emerge:
\begin{itemize}
\item While the strength of our proposed method is that it does not require any labeled samples, it might be of interest to study if and how the performance of the downstream models could be improved in case a small set of labeled samples is available which can be used when building $\transfer(\cdot)$.
\item Transparency is an important aspect of ML-based systems that are deployed in the real world. In this context, it would be of interest to make the adaptation $\transfer(\cdot)$ transparent, e.g. by explaining its outputs, which may provide some insights into the nature of the observed concept drift.
\item In this work, we always considered the scenario of a single concept drift. A straightforward extension to a scenario, where concept drifts occur frequently, is to completely retrain/rebuild $\transfer(\cdot)$ from scratch after every concept drift. In such scenarios, it would be of interest to explore incremental learning or adaptations of $\transfer(\cdot)$ in order to speed up the process of concept drift adaptation.
\end{itemize}

\section*{Acknowledgment}
A.A. and B.H. acknowledge funding from the VW-Foundation for the project \textit{IMPACT} funded in the frame of the funding line \textit{AI and its Implications for Future Society}, and funding from the European Research Council (ERC) under the ERC Synergy Grant Water-Futures (Grant agreement No. 951424).

K. M., C. G. P, and M. M. P. acknowledge funding from the European Research Council (ERC) under grant agreement No 951424 (Water-Futures), the European Union’s Horizon 2020 research and innovation programme under grant agreement No 739551 (KIOS CoE), and the Republic of Cyprus through the Deputy Ministry of Research, Innovation and Digital Policy.

\bibliographystyle{IEEEtran}
\bibliography{mybib}

\appendix

% \section{Experiments}\label{appendix:experiments}
Table~\ref{tab:appendix} shows the results on the Hanoi data set, where we report the median $R^2$ score over all scenarios, all numbers are rounded to two decimal points.

\begin{table}[h!]
\caption{Detailed results on the Hanoi data set}
\label{tab:appendix}
\centering
\footnotesize
\begin{tabular}{|c||c|c||c|c||}
 \hline
 Downstream task ID & $R^2\uparrow$ before drift & $R^2\uparrow$ after drift & $R^2\uparrow$ after drift \& AE reconstruction & $R^2\uparrow$ after \& drift adaptation  \\
 \hline\hline
 1 & $0.93$ & $0.85$ & $0.68$ & $0.87$ \\
2 & $0.93$ & $0.85$ & $0.68$ & $0.88$ \\
3 & $0.93$ & $0.85$ & $0.69$ & $0.88$ \\
4 & $0.93$ & $0.84$ & $0.69$ & $0.87$ \\
5 & $0.93$ & $0.84$ & $0.69$ & $0.87$ \\
6 & $0.93$ & $0.84$ & $0.69$ & $0.87$ \\
7 & $0.93$ & $0.84$ & $0.69$ & $0.87$ \\
8 & $0.93$ & $0.84$ & $0.69$ & $0.87$ \\
9 & $0.93$ & $0.84$ & $0.70$ & $0.87$ \\
10 & $0.93$ & $0.84$ & $0.69$ & $0.88$ \\
11 & $0.93$ & $0.84$ & $0.69$ & $0.87$ \\
12 & $0.93$ & $0.85$ & $0.70$ & $0.87$ \\
13 & $0.93$ & $0.86$ & $0.68$ & $0.87$ \\
14 & $0.93$ & $0.86$ & $0.68$ & $0.88$ \\
15 & $0.93$ & $0.86$ & $0.69$ & $0.88$ \\
16 & $0.93$ & $0.84$ & $0.70$ & $0.86$ \\
17 & $0.93$ & $0.85$ & $0.70$ & $0.86$ \\
18 & $0.93$ & $0.84$ & $0.69$ & $0.86$ \\
19 & $0.93$ & $0.85$ & $0.69$ & $0.88$ \\
20 & $0.93$ & $0.84$ & $0.69$ & $0.85$ \\
21 & $0.93$ & $0.83$ & $0.67$ & $0.85$ \\
22 & $0.93$ & $0.85$ & $0.69$ & $0.87$ \\
23 & $0.93$ & $0.86$ & $0.69$ & $0.88$ \\
24 & $0.93$ & $0.86$ & $0.69$ & $0.88$ \\
25 & $0.93$ & $0.86$ & $0.68$ & $0.88$ \\
26 & $0.93$ & $0.86$ & $0.68$ & $0.87$ \\
27 & $0.93$ & $0.86$ & $0.69$ & $0.88$ \\
28 & $0.93$ & $0.85$ & $0.69$ & $0.87$ \\
29 & $0.93$ & $0.83$ & $0.69$ & $0.87$ \\
30 & $0.93$ & $0.85$ & $0.69$ & $0.87$ \\
31 & $0.93$ & $0.85$ & $0.69$ & $0.87$ \\
 \hline
 \hline
\end{tabular}
\label{table:exp:results:hanoi}
\end{table}

\end{document}